\documentclass{article}

\usepackage[numbers]{natbib}
\usepackage[preprint]{neurips_2022}

\usepackage[dvipsnames]{xcolor}         
\definecolor{linkColor}{rgb}{0.18,0.39,0.62}
\usepackage[utf8]{inputenc} 
\usepackage[T1]{fontenc}    
\usepackage[colorlinks=true,linkcolor=linkColor,citecolor=linkColor,filecolor=linkColor,urlcolor=linkColor]{hyperref}       
\usepackage{url}            
\usepackage{booktabs}       
\usepackage{amsfonts}       
\usepackage{nicefrac}       
\usepackage{microtype}      

\usepackage{graphicx}
\usepackage{arydshln}
\usepackage{booktabs}
\usepackage{multirow}
\usepackage{caption}
\usepackage{subcaption}
\usepackage{makecell}
\usepackage{csquotes}
\usepackage{epigraph}
\usepackage{amsmath}
\usepackage{xcolor}  
\usepackage{cleveref}
\usepackage{pifont}
\usepackage{listings}

\RequirePackage{algorithm}
\RequirePackage{algorithmic}

\usepackage{multirow}
\usepackage{amsmath}
\usepackage{capt-of}
\usepackage{tabularx}
\usepackage{epsfig}
\usepackage{amssymb}
\usepackage{amsfonts}
\usepackage{booktabs}
\usepackage{scalerel}
\usepackage[inline]{enumitem}
\usepackage{listings}
\usepackage{varwidth}
\usepackage[export]{adjustbox}
\usepackage{tikz}
\usetikzlibrary{tikzmark}

\usepackage{stmaryrd}
\usepackage{bbm}
\usepackage{wrapfig}
\usepackage{pifont}

\definecolor{deepblue}{rgb}{0,0,0.5}
\definecolor{officeblue}{RGB}{0,102,204}
\definecolor{deepred}{rgb}{0.6,0,0}
\definecolor{deepgreen}{rgb}{0,0.5,0}
\definecolor{mybrickred}{RGB}{182,50,28}

\definecolor{fillcolor}{RGB}{216,217,252}

\usepackage{etoolbox}
\usepackage{framed}

\newif\ifxetexorluatex
\ifxetex
  \xetexorluatextrue
\else
  \ifluatex
    \xetexorluatextrue
  \else
    \xetexorluatexfalse
  \fi
\fi
\newcommand*\quotesize{60} 
\newcommand*{\openquote}
   {\tikz[remember picture,overlay,xshift=-4ex,yshift=-2.5ex]
   \node (OQ) {\fontsize{\quotesize}{\quotesize}\selectfont``};\kern0pt}

\newcommand*{\closequote}[1]
  {\tikz[remember picture,overlay,xshift=4ex,yshift={#1}]
   \node (CQ) {\fontsize{\quotesize}{\quotesize}\selectfont''};}

\colorlet{shadecolor}{white}

\newcommand*\shadedauthorformat{\emph} 

\newcommand*\authoralign[1]{%
  \if#1l
    \def\authorfill{}\def\quotefill{\hfill}
  \else
    \if#1r
      \def\authorfill{\hfill}\def\quotefill{}
    \else
      \if#1c
        \gdef\authorfill{\hfill}\def\quotefill{\hfill}
      \else\typeout{Invalid option}
      \fi
    \fi
  \fi}
{\authoralign{#1}
\ifblank{#2}
   {\def\shadequoteauthor{}\def\yshift{-2ex}\def\quotefill{\hfill}}
   {\def\shadequoteauthor{\par\authorfill\shadedauthorformat{#2}}\def\yshift{2ex}}
\begin{snugshade}\begin{quote}\openquote}
{\shadequoteauthor\quotefill\closequote{\yshift}\end{quote}\end{snugshade}}

\usepackage{amsmath,amsfonts,bm}

\def\eqref#1{equation~\ref{#1}}

\def\1{\bm{1}}

\DeclareMathAlphabet{\mathsfit}{\encodingdefault}{\sfdefault}{m}{sl}
\SetMathAlphabet{\mathsfit}{bold}{\encodingdefault}{\sfdefault}{bx}{n}

\newcommand\our{\text{BitNet a4.8}}
\newcommand\bitnet{\text{BitNet b1.58}}
\newcommand\llama{\text{LLaMA LLM}}

\title{BitNet a4.8: 4-bit Activations for 1-bit LLMs}

\author{
 Hongyu Wang\thanks{~Equal contribution. $\diamond$ Corresponding author. S. Ma and F. Wei are with Microsoft Research. H. Wang is with University of Chinese Academy of Sciences.}~~~~Shuming Ma\footnotemark[1]~~~~Furu Wei$^{\diamond}$ \\ \\
{\href{https://aka.ms/GeneralAI}{https://aka.ms/GeneralAI}}
\\}

\begin{document}
\maketitle
\begin{abstract}
Recent research on the 1-bit Large Language Models (LLMs), such as BitNet b1.58~\cite{bitnet2}, presents a promising direction for reducing the inference cost of LLMs while maintaining their performance. In this work, we introduce \textbf{\our{}}, enabling 4-bit activations for 1-bit LLMs. \our{} employs a hybrid quantization and sparsification strategy to mitigate the quantization errors introduced by the outlier channels. Specifically, we utilize 4-bit activations for inputs to the attention and feed-forward network layers, while sparsifying intermediate states followed with 8-bit quantization. Extensive experiments demonstrate that \our{} achieves performance comparable to BitNet b1.58 with equivalent training costs, while being faster in inference with enabling 4-bit (INT4/FP4) kernels. Additionally, \our{} activates only 55\% of parameters and supports 3-bit KV cache, further enhancing the efficiency of large-scale LLM deployment and inference.
\end{abstract}

\vspace{-0.6cm}

\begin{figure}[h]
    \centering
    \includegraphics[width=\textwidth]{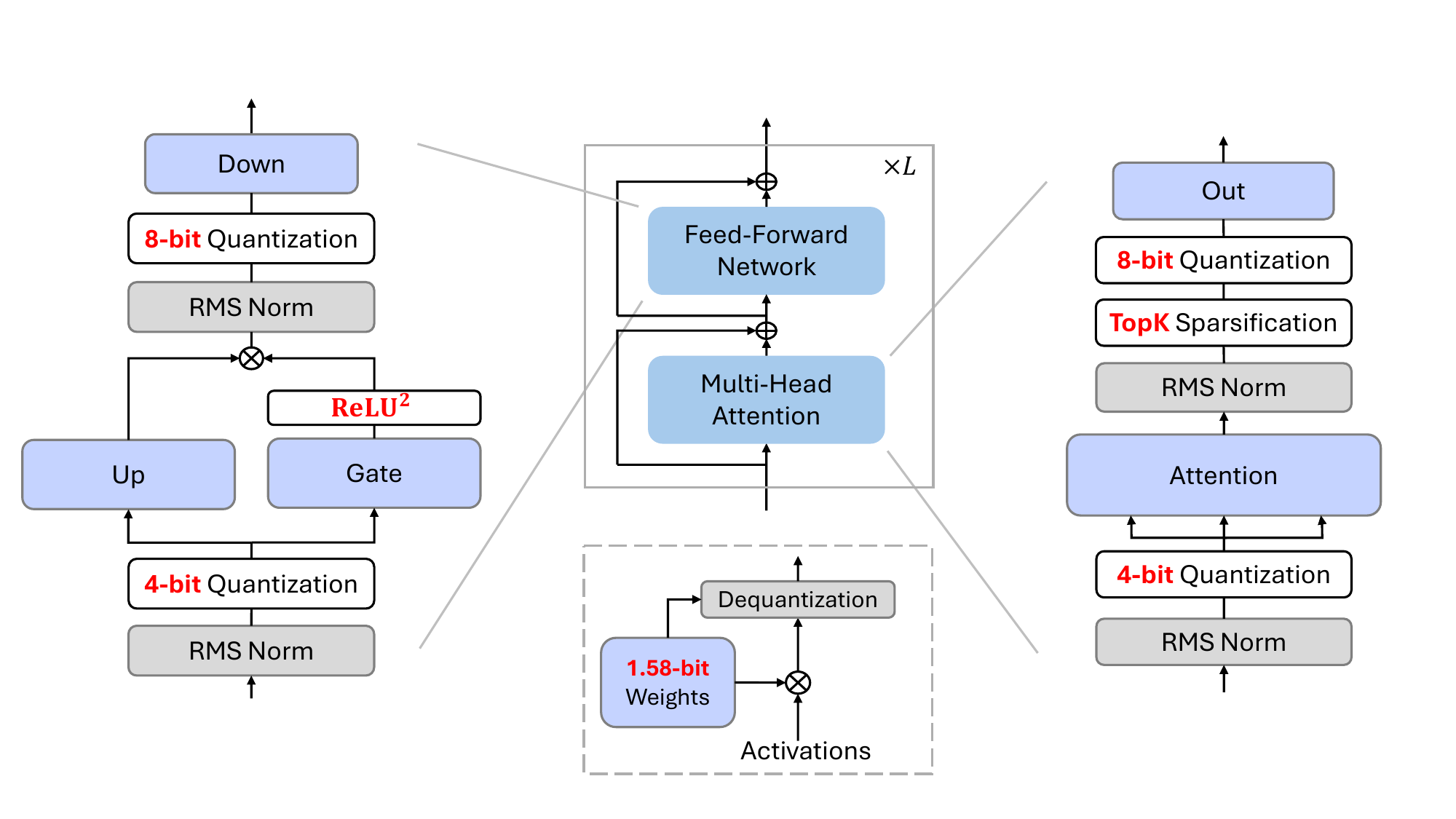}
    \caption{The overview of \our{} with both weight and activation quantization. All the parameters are ternery (i.e., 1.58-bit as in BitNet b1.58~\cite{bitnet2}). We use a hybrid quantization
and sparsification strategy to deal with outlier activations in certain Transformer sub-layers.}
    \label{fig:overview}
\end{figure}

\newpage
\section{Introduction}
Recent works~\cite{bitnet2} have shown that 1-bit LLMs can match the full-precision models given the same amount of parameters and training tokens while being significantly cost-effective in terms of latency, memory, throughput, and energy consumption. With model weights represented in 1.58-bit (i.e., \{-1, 0, 1\}), the bottleneck of inference has shifted from the limited memory bandwidth to high computational cost. Low-bit or sparse activations in LLMs served as a promising approach to further reduce the computational budget while maintaining performance on downstream tasks.

One common approach is to utilize activation sparsity~\cite{dejavu, turbosparse, teal}, which reduces the inference FLOPs and the I/O of weight by pruning the activation entries with smaller magnitudes. Sparsification is particularly well-suited for handling activations that exhibit highly imbalanced long-tailed distributions. Recent works~\cite{qsparse} have demonstrated that LLMs with fully sparsely-activated activations can achieve results comparable to dense models while having much less active parameters.

In addition to sparsification, activation quantization is another approach to accelerate the matrix multiplication. However, the optimization of neural networks with low-bit activations is challenging due to the emergence of outlier dimensions as the training progresses and the model size grows. Despite these outliers only account for a very small portion of the activations~\cite{llmint8, smoothquant}, they have much larger magnitude, which leads to significant quantization errors and performance degradation on downstream tasks. Previous works~\cite{int4, quarot, spinquant, duquant} mostly utilize Hadamard or learnable rotary transformation to amortize the outlier features into other entries. However, they are mostly designed for the LLMs of higher precision (e.g., 4-bit). For 1-bit LLMs, the extremely low bit-width of the weights makes it challenging to absorb these transformation matrices directly into the weights, while leaving them as online transformations introduces additional computational overhead and limits overall inference performance.

In this work, we introduce \textbf{\our{}}, a hybrid quantization and sparsification strategy that enables 4-bit activations for 1-bit LLMs. By carefully analyzing the activation distribution of 1-bit LLMs, we selectively apply 4-bit quantization or sparsification based on the distribution patterns of these activations. Specifically, as shown in Figure~\ref{fig:overview}, \our{} employs 4-bit activations for the inputs to attention and FFN, while utilizing sparsification with 8 bits for intermediate states. To improve the training efficiency, \our{} is trained from 8-bit to 4-bit activations with a two-stage recipe, which requires only a few training tokens to adapt \bitnet{} to the low-bit activations at the end of training. Extensive experiments demonstrate that \our{} achieves competitive performance to \bitnet{} with the same training cost while being significantly more efficient at inference time. Furthermore, \our{} has only 55\% activated parameters and supports 3-bit KV cache, which further enhances the efficiency of LLM deployment.

\section{\our{}}

\subsection{Architecture}
\label{sec:arch}

As shown in Figure~\ref{fig:overview}, \our{} adopts the same layout as \bitnet{}. Following~\cite{bitnet, bitnet2}, we replace the linear projections in both attention and feed-forward network (FFN) with BitLinear to learn 1.58-bit weights from the scratch. For activations, we adopt a hybrid quantization and sparsification strategy to mitigate the error introduced by outlier dimensions. 

\begin{figure}[t]
    \centering
    \includegraphics[width=\textwidth]{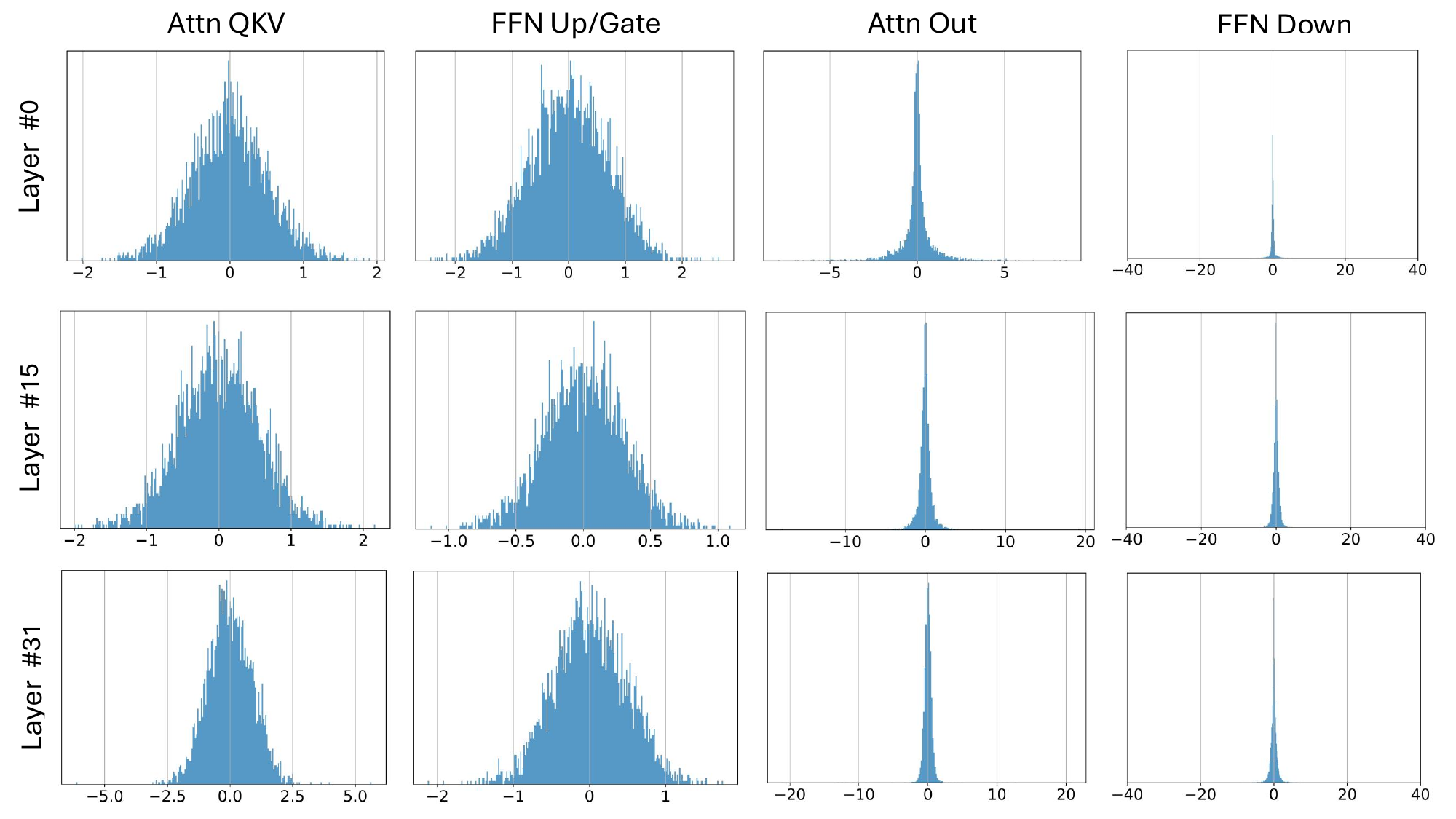}
    \caption{The distribution of the inputs to each projection. The visualization is conducted with a 7B BitNet b1.58 model on a subset of the valid set of C4. For the layers that exhibit Gaussian-like distributions, we employ 4-bit activation quantization. For the layers which distributions are sharp, we adopt Q-Sparse~\cite{qsparse} to perform sparsification on the activations.}
    \label{fig:act}
\end{figure}

\begin{figure}[t]
    \centering
    \includegraphics[width=\textwidth]{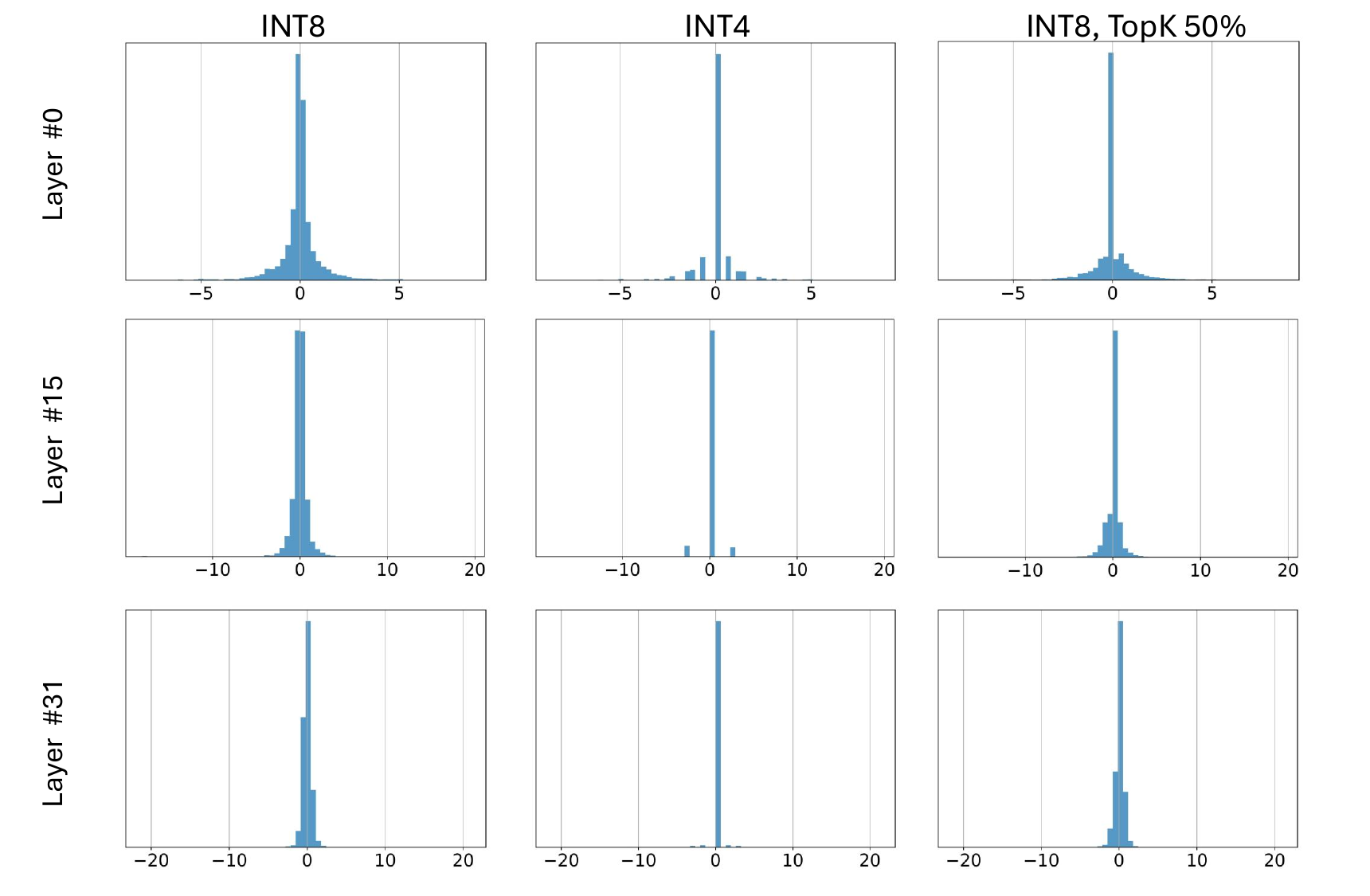}
    \caption{The distribution of the inputs to the output projection of attention with different quantization and sparsification. The visualization is conducted with a 7B BitNet b1.58 model on a subset of the valid set of C4.}
    \label{fig:outproj_act_dist}
\end{figure}

Figure~\ref{fig:act} illustrates the distribution of each component's inputs of a BitNet b1.58 model with 7B model size. Inputs to the attention and FFN layers typically follow a Gaussian-like distribution, while activations before the FFN down projections and the output projections in attention have more outlier channels and massive amount of entries around zero. \cite{teal} also reported similar observations for full-precision LLMs. As shown in Figure~\ref{fig:outproj_act_dist}, directly applying low-bit quantization to these intermediate states introduces substantial quantization errors.

Therefore, we use the sparsification method from Q-Sparse~\cite{qsparse} to retain these intermediate states at 8 bits while removing the computation bottleneck. For output projection of self-attention layers, we use a sparsify-then-quantize function:

\begin{equation}
    \mathbf{Y} = (\text{Q}_{\text{INT8}}(\mathbf{X}) \odot \mathbf{M}) \cdot \text{Q}_{w}(\mathbf{W})^T, \,\mathbf{M} = \text{Top}_k (\mathbf{|X|})
\end{equation}

where $\text{Q}_{w}(\cdot)$ and $\text{Q}_{\text{INT8}}(\cdot)$ denote the quantization function for weight $\mathbf{W}$ and activations $\mathbf{X}$, respectively. $\mathbf{M}$ is the mask tensor that indicates the maximum top-K elements in terms of the absolute values of the activations $\mathbf{X}$, $\odot$ is the element-wise multiplication operation.

Specifically, the functions of weight quantization and activation quantization can be formulated as:

\begin{align}
    &\text{Q}_{w}(\mathbf{W}) = \alpha\text{RoundClip}(\frac{\mathbf{W}}{\alpha+\epsilon}, -1, 1),\,\alpha = \text{mean}(|\mathbf{W}|) \\
    &\text{Q}_{\text{INT8}}(\mathbf{X}) = \frac{\gamma}{127}\text{RoundClip}(\frac{127}{\gamma+\epsilon}\mathbf{X}, -128, 127),\,\gamma = \max(|\mathbf{X}|) \\
    &\text{RoundClip}(X, a, b) = \min(\max(\text{round}(X), a), b)
\end{align}

For FFN, we adopt squared ReLU~\cite{primer, qsparse} and gated linear unit (GLU) to further boost the activation sparsity. It is defined as follows:

\begin{equation}
    \text{ReLU}^2\text{GLU}(\mathbf{X}) = \mathbf{X} \mathbf{W}_{\text{up}}^{T} \odot \text{ReLU}^2(\mathbf{X} \mathbf{W}_{\text{gate}}^T)
\end{equation}

According to our preliminary experiments, with squared ReLU, the inputs to the down projection achieve over 80\% sparsity with minimal impact on performance. Additionally, we observe that the outputs of gate projection $\text{ReLU}^2(\mathbf{X} \mathbf{W}_{\text{gate}}^T)$ exhibit high activation sparsity as well (e.g., 67.5\% for 7B models). This characteristic enables further reduction in inference FLOPs for the up projection by first computing the gate projection and then performing the up projection only on the non-zero channels of the gates.

For the input to attention and FFN, since they have much less outlier features, we use absmean function to quantize the activations to 4-bit integers:
\begin{align}
    &\mathbf{Y} = \text{Q}_{\text{INT4}}(\mathbf{X}) \cdot \text{Q}_{w}(\mathbf{W})^T \\
    &\text{Q}_{\text{INT4}}(\mathbf{X}) = \frac{\beta}{\sqrt{7}}\text{RoundClip}(\frac{\sqrt{7}}{\beta+\epsilon}\mathbf{X}, -8, 7),\,\beta = \text{mean}(|\mathbf{X}|)
\end{align}

\subsection{Training}

\paragraph{Continue-training from \bitnet{}.} \our{} is trained with a two-stage recipe from W1.58A8 to W1.58A4. For the first stage, we train the model with 8-bit activations and $\text{ReLU}^2\text{GLU}$. For the second stage, we adopt the hybrid quantization and sparsification as shown in Section~\ref{sec:arch}. \our{} quickly adapts to 4-bit and sparse activations with only a few training tokens while having negligible loss on performance.

\paragraph{Gradient approximation.} Following~\cite{bitnet, qsparse}, we use straight-through estimator (STE)~\cite{ste} to conduct the gradient approximation for \our{}, as well as mixed precision training to update the parameters. We directly bypass the non-differentiable functions, including the quantization function and top-K sparsification function during the backward propagation. For mixed precision training, we maintain a full-precision latent weight to accumulate parameter updates. During the forward, we quantize the latent weight into 1.58-bit on the fly.

\subsection{Floating-point quantization}

Floating-point quantization offers a broader dynamic range than the integer-based quantization, which is crucial for handling the long-tailed distribution of the activations. For floating-point precision, we only leave the inputs to down projection of FFN at 8-bit integers, and quantize the other activations to FP4 using MinMax quantizer~\cite{llm-fp4}. It is defined as follows:
\begin{align}
    &\text{Q}_{\text{FP4}}(\mathbf{X}) = \frac{\gamma}{2^{M+b}}\text{Round}(\frac{2^{M+b}}{\gamma}\mathbf{X}),\,\,\gamma = 2^{\text{max}(\lfloor \lfloor \log_2|\mathbf{X}| \rfloor + b\rfloor, 1)} \\
    &b = \log_2(\cfrac{2 - 2^{-M}}{|\mathbf{X}|_{\text{max}}}) + 2^E - 1
\end{align}

where $E$ and $M$ denote the bit-width of the exponent and mantissa component, respectively. We adopt the E2M1 format due to its larger dynamic range. As shown in Table~\ref{tab:zero-shot}, \our{} with FP4 quantization has the similar performance as it with the hybrid quantization and sparsification strategy based on integers.

\section{Experiments}

We compared \our{} to \bitnet{} and our reproduced FP16 LLaMA LLM of various sizes. For 1.58-bit models, we adopted the two-stage weight decay and learning rate scheduling following the training recipe of \bitnet{}~\cite{bitnet2}. More details can be found in the Appendix~\ref{ap:hyper}. All models were trained with 100B tokens from the RedPajama dataset~\cite{redpajama} to ensure a fair comparison. For \our{}, we first train the model with 8-bit activations for 95B tokens. Then we reuse the optimizer states and continue-train the model with the proposed hybrid quantization and sparsification for 5B tokens. We set topK as 50\% for the output projection of attention. 

We evaluated the zero-shot accuracy for these models on a range of language tasks using the \emph{lm-evaluation-harness} toolkit~\citep{eval-harness}, including ARC-Easy (ARCe)~\cite{arc}, ARC-Challenge (ARCc)~\cite{arc}, Hellaswag (HS)~\cite{hellaswag}, Winogrande (WGe)~\cite{winoGrande} and PIQA (PQ)~\cite{piqa}. We also reported the perplexity on the validation set of C4~\cite{c4} dataset.

\subsection{Main Results}

Table~\ref{tab:zero-shot} summarizes the detailed results of \our{}, BitNet b1.58 and FP16 \llama{}. The performance gap between full-precision (i.e., FP16) \llama{} and BitNet b1.58 narrows as the model size grows. For 7B models, BitNet b1.58 matches \llama{} in terms of both language model perplexity and average accuracy on the end tasks. Furthermore, \our{} achieves performance comparable to BitNet b1.58, with almost no loss in average accuracy.

\begin{table*}[t]
    \centering
    \begin{tabular}{lcccccccc}
    \toprule
    \textbf{Models} & \textbf{Size} & \textbf{PPL$\downarrow$} & \textbf{ARCc$\uparrow$} & \textbf{ARCe$\uparrow$} & \textbf{HS$\uparrow$} & \textbf{PQ$\uparrow$} & \textbf{WGe$\uparrow$} & \textbf{Avg$\uparrow$} \\
    \midrule
    \llama & \multirow{4}{*}{700M} & 11.44 & 27.13 & 43.27 & 44.70 & 68.12 & 53.99 & 47.44\\ 
    \bitnet &  & 12.32 & 25.00 & 42.68 & 42.08 & 66.97 & 54.14 & 46.17 \\
    \textbf{\our{}} (FP4) & & 12.40 & 25.17 & 42.68 & 42.36 & 66.27 & 52.96 & 45.89\\
    \textbf{\our{}} & & 12.40 & 25.17 & 41.58 & 42.44 & 66.38 & 53.04 & 45.72\\
    \midrule
    \llama & \multirow{4}{*}{1.3B} & 10.82 & 27.90 & 45.16 & 47.65 & 69.91 & 53.35 & 48.79 \\
    \bitnet & & 11.27 & 27.65 & 45.33 & 46.86 & 68.39 & 54.06 & 48.46\\
    \textbf{\our{}} (FP4) & & 11.38 & 28.50 & 44.36 & 47.03 & 68.61 & 54.06 & 48.51 \\
    \textbf{\our{}} & & 11.35 & 28.50 & 44.15 & 46.98 & 68.34 & 54.14 & 48.42\\
    \midrule
    \llama & \multirow{4}{*}{3B} & 9.61 & 29.95 & 48.11 & 55.25 & 71.76 & 57.46 & 52.51\\
    \bitnet & & 9.97 & 29.27 & 49.41 & 54.42 & 70.89 & 57.54 & 52.30\\
    \textbf{\our{}} (FP4) & & 9.99 & 29.10 & 49.24 & 54.60 & 71.38 & 56.12 & 52.08 \\
    \textbf{\our{}} & & 9.97 & 28.33 & 49.58 & 54.62 & 71.16 & 54.38 & 51.61\\
    \midrule
    \llama & \multirow{4}{*}{7B} & 9.20 & 33.36 & 51.22 & 58.33 & 73.34 & 58.41 & 54.93 \\
    \bitnet & & 9.24 & 32.00 & 50.88 & 59.79 & 72.96 & 59.83 & 55.09 \\
    \textbf{\our{}} (FP4) & & 9.42 & 31.57 & 51.22 & 58.20 & 72.47 & 59.59 & 54.61\\
    \textbf{\our{}} & & 9.37 & 31.66 & 50.88 & 58.78 & 73.01 & 59.35 & 54.74\\
    \bottomrule
    \end{tabular}
    \caption{Perplexity and results of \our{}, BitNet b1.58 and \llama{} on the end tasks. The standard variance of error for average scores is 1.06\%.}
    \label{tab:zero-shot}
\end{table*}

\begin{table*}[t]
    \centering
    \begin{tabular}{lccccccc}
    \toprule
    \textbf{Models} & \textbf{Activated} & \textbf{QKV} & \textbf{Out} & \textbf{Up} & \textbf{Gate} & \textbf{Down} & \textbf{Overall} \\
    \midrule
    \llama & 679M & 0.0 & 0.0 & 0.0 & 0.0 & 0.0 & 0.0 \\
    BitNet b1.58 & 638M & 1.2 & 5.9 & 1.2 & 1.2 & 21.8 & 6.2 \\
    \bf \our{} & 390M & 12.1 & 50.0 & 66.2 & 12.1 & 80.9 & 42.5 \\
    \midrule
    \llama & 1.2B & 0.0 & 0.0 & 0.0 & 0.0 & 0.0 & 0.0 \\
    BitNet b1.58 & 1.1B & 1.3 & 5.8 & 1.2 & 1.2 & 22.8 & 6.4 \\
    \bf \our{} & 0.7B & 12.0 & 50.0 & 65.9 & 12.1 & 81.8 & 42.7 \\
    \midrule
    \llama & 3.2B & 0.0 & 0.0 & 0.0 & 0.0 & 0.0 & 0.0 \\
    BitNet b1.58 & 3.0B & 1.4 & 7.1 & 1.3 & 1.3 & 30.0 & 8.2 \\
    \bf \our{} & 1.8B & 12.1 & 50.0 & 70.7 & 12.1 & 85.6 & 44.7 \\
    \midrule
    \llama & 6.5B & 0.0 & 0.0 & 0.0 & 0.0 & 0.0 & 0.0 \\
    BitNet b1.58 & 6.0B & 1.7 & 11.2 & 1.4 & 1.4 & 24.2 & 7.3\\
    \bf \our{} & 3.4B & 12.1 & 50.0 & 71.4 & 12.0 & 84.2 & 44.5 \\
    \bottomrule
    \end{tabular}
    \caption{Detailed sparsity of \our{}, BitNet b1.58 and \llama{} on the valid set of C4.}
    \label{tab:sparsity}
\end{table*}

\paragraph{Sparsity.} Table~\ref{tab:sparsity} demonstrates detailed sparsity of each component for \our{}, BitNet b1.58 and FP16 \llama{} across various sizes. The sparsity is calculated with non-embedding parameters on the valid set of C4. Notably, \our{} achieves significantly higher sparsity than both BitNet b1.58 and \llama{}. For example, in the 7B model, \our{} reaches an overall sparsity of 44.5\%, with only 3.4B active parameters. The inputs to the down projection demonstrate particularly high sparsity, consistent with our observation that intermediate state distributions are sharply centered around zero. Additionally, we also observe that the outputs of gate projection are very sparse. It leads to a high sparsity for up projection, since we only need to perform projection on the non-zero channels selected from the gates. Specifically, for the 7B \our{}, the sparsity of the gates and the inputs to up projection is 67.5\% and 12.0\%, respectively. Consequently, the sparsity of up projection can be estimated as $1 - (1 - 12.0\%)\times(1 - 67.5\%)$, that is 71.4\%.

\begin{table*}[t]
    \centering
    \begin{tabular}{lccccccc}
    \toprule
    \textbf{Models} & \textbf{Size} & \textbf{ARCc$\uparrow$} & \textbf{ARCe$\uparrow$} & \textbf{HS$\uparrow$} & \textbf{PQ$\uparrow$} & \textbf{WGe$\uparrow$} & \textbf{Avg$\uparrow$} \\
    \midrule
    \our{} & \multirow{4}{*}{3B} &  28.33 & 49.58 & 54.62 & 71.16 & 54.38 & 51.61\\
    w/ 4-bit KV & & 28.24 & 48.86 & 54.41 & 71.87 & 55.49 & 51.77 \\
    w/ 4-bit QKV & & 27.30 & 48.91 & 54.32 & 71.98 & 56.75 & 51.85\\
    w/ 4-bit Q, 3-bit KV & & 28.84 & 48.91 & 53.87 & 70.95 & 56.35 & 51.78 \\
    \midrule
    \our{} & \multirow{4}{*}{7B} & 31.66 & 50.88 & 58.78 & 73.01 & 59.35 & 54.74\\
    w/ 4-bit KV & & 31.40 & 50.93 & 58.68 & 73.12 & 60.85 & 55.00\\
    w/ 4-bit QKV & & 30.63 & 51.30 & 58.45 & 72.52 & 59.83 & 54.55\\
    w/ 4-bit Q, 3-bit KV & & 31.14 & 50.93 & 58.07 & 72.96 & 59.04 & 54.43 \\
    \bottomrule
    \end{tabular}
    \caption{Detailed results of \our{} with QKV states varying bit-widths on the end tasks. We reported the zero-shot accuracy of all models.}
    \label{tab:attn}
\end{table*}

\paragraph{Low-bit Attention.} Table~\ref{tab:attn} presented detailed results of \our{} with low-bit attention in 3B and 7B model size. Low-bit attention is essential for efficient long sequence modeling, as it reduces the memory footprint and IO of KV cache and accelerates the attention computation. In our experiments, we adopted post-RoPE quantization. The QKV heads were directly quantized to unsigned integers using the absmax function, without the need of any calibration dataset. For 3-bit KV quantization, we retain the heads of the bos token at 4-bit, as it contains more outlier features. As shown in Table~\ref{tab:attn}, \our{} achieves negligible accuracy loss with 4-bit KV or QKV heads in 3B and 7B models. Furthermore, the KV cache of \our{} can be quantized to 3-bit integers, resulting in almost no degradation on average accuracy.

\subsection{Ablation Study}

\begin{figure*}[t]
    \centering
    \begin{subfigure}{0.45\textwidth}
        \centering
        \includegraphics[width=\linewidth]
        {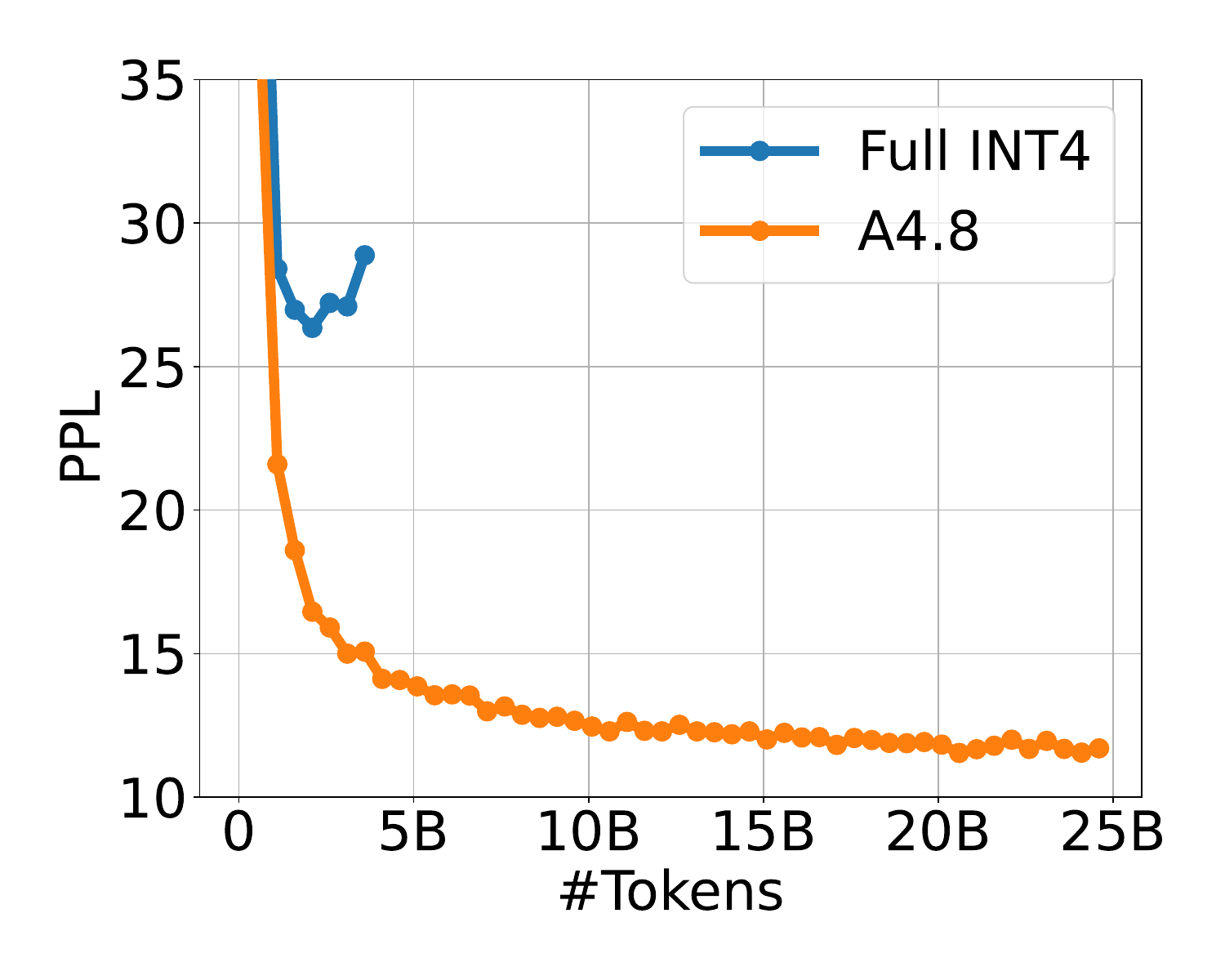}
    \end{subfigure}
    \begin{subfigure}{0.45\textwidth}
        \centering
        \includegraphics[width=\linewidth]
        {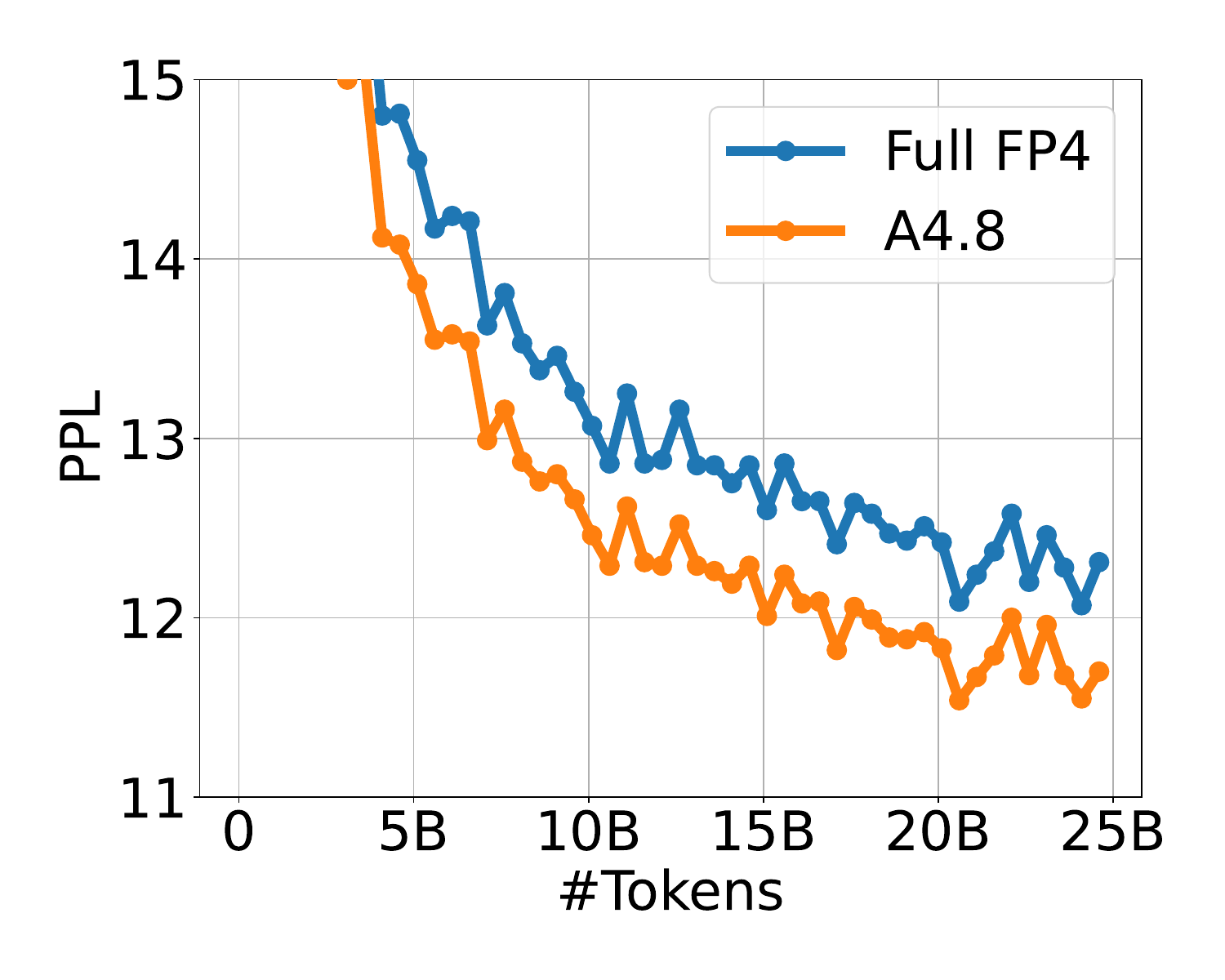}
    \end{subfigure}
    \caption{Ablation study on the hybrid quantization and sparsification.}
    \label{fig:hybrid}
\end{figure*}

\begin{figure*}[t]
    \centering
    \begin{subfigure}{0.45\textwidth}
        \centering
        \includegraphics[width=\linewidth]
        {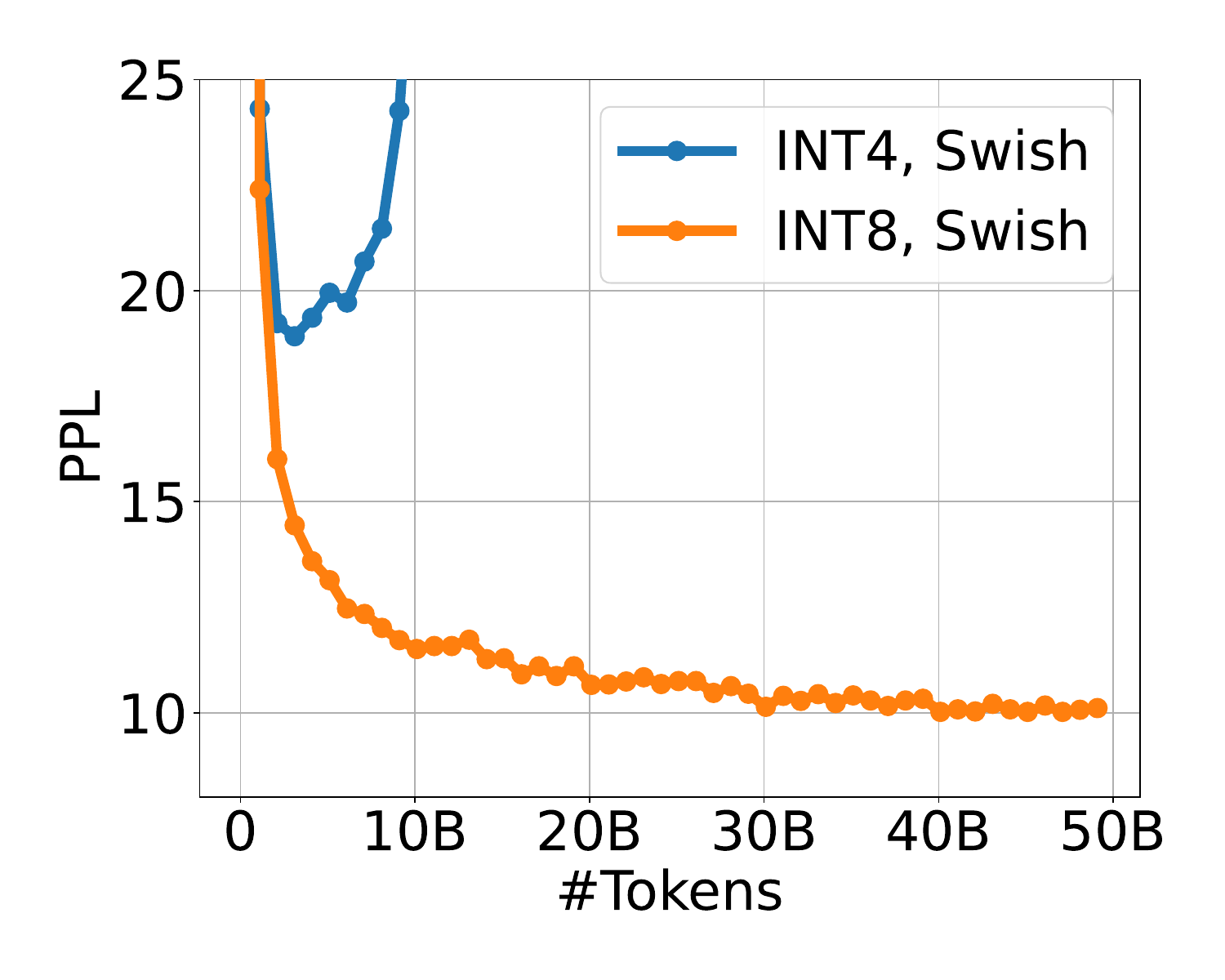}
    \end{subfigure}
    \begin{subfigure}{0.45\textwidth}
        \centering
        \includegraphics[width=\linewidth]
        {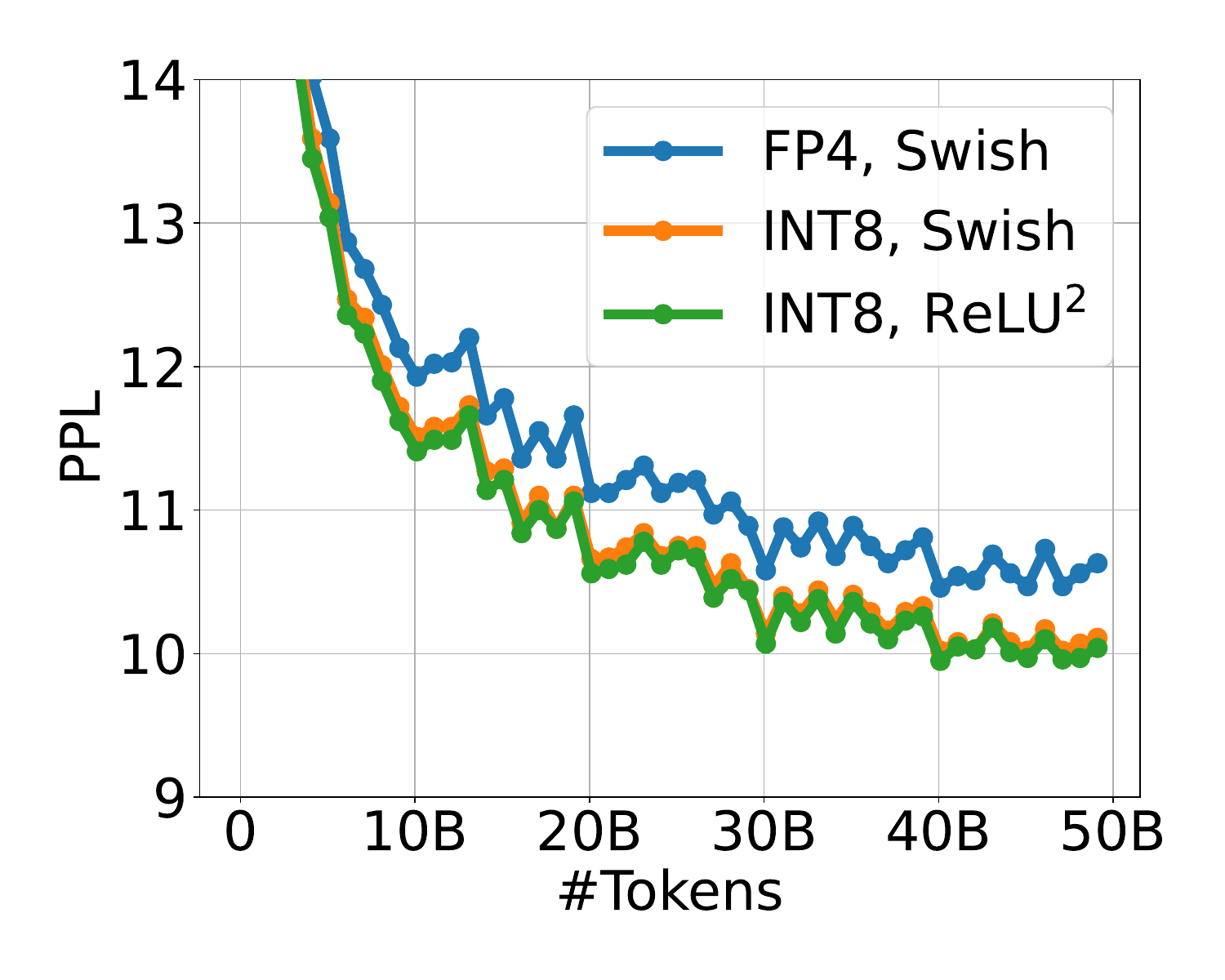}
    \end{subfigure}
    \caption{Ablation study on different quantization or activation function for the inputs to down projection of FFN.}
    \label{fig:down}
\end{figure*}

\begin{table*}[t]
    \centering
    \scalebox{0.98}{
    \begin{tabular}{cc|ccccccc}
    \toprule
    \textbf{Quantization} & \textbf{Sparsification} & \textbf{PPL$\downarrow$} & \textbf{ARCc$\uparrow$} & \textbf{ARCe$\uparrow$} & \textbf{HS$\uparrow$} & \textbf{PQ$\uparrow$} & \textbf{WGe$\uparrow$} & \textbf{Avg$\uparrow$} \\
    \midrule
    INT8 & - & 9.95 & 28.33 & 48.53 & 54.90 & 72.31 & 56.51 & 52.11 \\
    INT8 & TopK 50\% & 9.97 & 28.33 & 49.58 & 54.62 & 71.16 & 54.38 & 51.61\\
    \bottomrule
    \end{tabular}
    }
    \caption{Ablations on the TopK sparsification for the inputs to the output projection of attention.}
    \label{tab:ablation_outproj}
\end{table*}

\paragraph{Hybrid architecture.} Figure~\ref{fig:hybrid} presented the training loss curve of 700M \our{} with the full INT4/FP4 quantization, and the hybrid quantization and sparsification. We train these models with the first-stage scheduling for 25B tokens from the RedPajama dataset. We adopt absmean and MinMax quantizer for full INT4 and FP4 quantization, respectively. Besides, for full INT4 quantization, we use absmean quantizer with $\beta = 2 \text{mean}(|X|)$ for down projection in FFN, as its inputs have larger outliers. As shown in Figure~\ref{fig:hybrid}, the full INT4 quantization leads to divergence. Furthermore, the hybrid architecture significantly outperforms the full FP4 architecture in terms of training perplexity.

\paragraph{Down projection of FFN.} We compared 1.3B \our{} with different quantization or activation function for the down projection of FFN. All models were trained with the first-stage scheduling for 50B tokens from the RedPajama dataset. To ensure a fair comparison, we leave the other activations at 8-bits. We adopt the absmax quantizer for INT8 quantization and MinMax quantizer for FP4 quantization. The $\beta$ of absmean quantizer is set as $2 \text{mean}(|X|)$. Figure~\ref{fig:down} shows the training loss curves of these models. Squared ReLU achieves slightly better training perplexity than Swish while enabling higher sparsity. Furthermore, applying FP4 quantization for the inputs to the down projection leads to a significant performance degradation, while using INT4 activations with STE causes divergence.

\begin{wrapfigure}{r}{5cm}
    \centering
    \includegraphics[width=\linewidth]{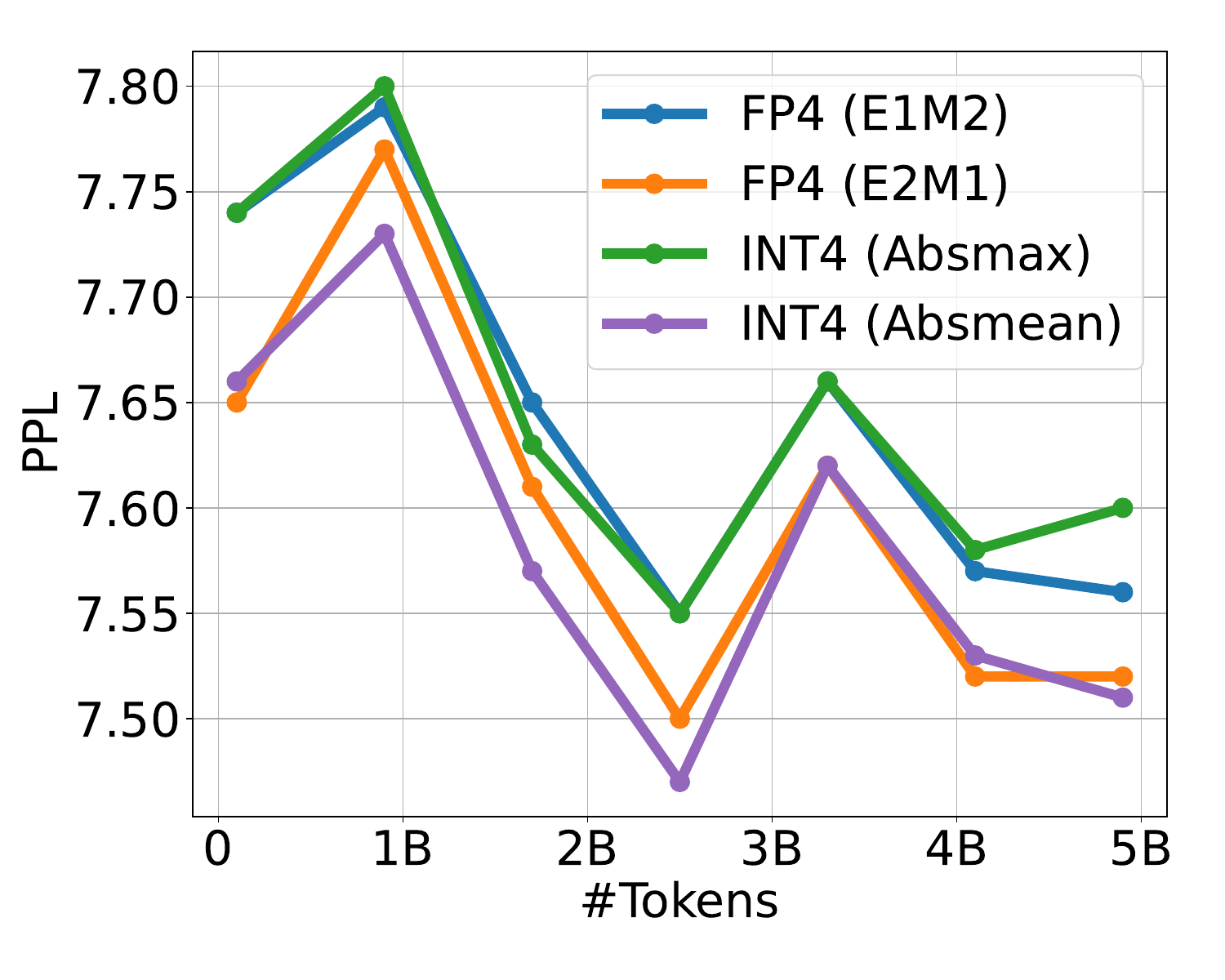}
    \caption{Ablations on 4-bit quantizers for the inputs to attention and FFN.}
    \label{fig:4bit}
\end{wrapfigure}

\paragraph{Output projection of attention.} Table~\ref{tab:ablation_outproj} demonstrates detailed results of 3B \our{} with and without Top-K sparsification for the inputs to the output projection of attention. Both models were trained with the same two stage recipe from 8-bit to 4-bit activations. We set $K$ as 50\% for sparsification. The baseline utilized the INT8 absmax quantizer for the output projection's inputs. The results show that TopK sparsification brings negligible perplexity and accuracy loss.

\paragraph{4-bit quantization.} We presented the loss curves of 3B \our{} with different 4-bit quantizers for the inputs to the attention and FFN. We compared the performance of \our{} with floating-point quantization with E2M1 and E1M2 formats using MinMax quantizer, integer quantization with absmax and absmean quantizer. As shown in Figure~\ref{fig:4bit}, FP4 with E2M1 format and INT4 with absmean quantizer achieve slightly better training perplexity, as they are well-suited for handling small-magnitude activation entries.

\begin{table*}[t]
    \centering
    \scalebox{0.9}{
    \begin{tabular}{lccccccccc}
    \toprule
    \textbf{Models} & \textbf{HS} & \textbf{PQ} & \textbf{WGe} & \textbf{OBQA} & \textbf{Lambada} & \textbf{MMLU} & \textbf{ARCc} & \textbf{ARCe} & \textbf{Avg} \\
    \midrule
    \bitnet{} 2B & 68.66 & 77.09 & 62.58 & 41.40 & 63.36 & 50.29 & 47.61 & 70.74 & 60.22 \\
    \textbf{\our{} 2B} & 68.21 & 76.55 & 64.40 & 40.60 & 63.75 & 50.30 & 46.59 & 70.00 & 60.05 \\
    \bottomrule
    \end{tabular}
    }
    \caption{Results of \our{} and \bitnet{} with 2B parameters and 2T training tokens.}
    \label{2b2t}
\end{table*}

\subsection{More Training Tokens}

Prior research~\cite{llmint8} demonstrates a positive correlation between the number of training tokens and the prevalence of activation outliers in language models. To rigorously evaluate the scalability characteristics of \our{}, we conducted extensive experiments using a model configuration with 2 billion parameters trained on 2 trillion tokens. We performed a controlled comparison against \bitnet{} using identical training data and configurations. The empirical results, presented in Table~\ref{2b2t}, demonstrate that \our{} maintains performance parity with negligible degradation in accuracy metrics while achieving 4-bit activation compression. These findings provide strong evidence for the efficacy of our proposed approach at scale.

\section{Conclusion}
In this paper, we present \our{} which enables 4-bit activations for 1-bit LLMs. \our{} uses a novel hybrid quantization and sparsification architecture to reduce the quantization errors introduce by outlier channels of activations. Specifically, we employ 4-bit quantization for inputs to the attention and FFN layers, while sparsifying the intermediate states with 8-bit integers. \our{} is continue-trained from W1.58A8 to W1.58A4. Experimental results demonstrate that \our{} achieves results comparable to BitNet b1.58 with the same training cost, while significantly enhancing inference efficiency. 

\section{Acknowledgements}
We would like to acknowledge Lei Wang for the discussion on the inference efficiency.

\bibliography{bitnet}
\bibliographystyle{alpha}

\appendix

\section{Hyper-parameters}
\label{ap:hyper}

\begin{table*}[h]
\setlength{\tabcolsep}{5pt}
\centering
\begin{tabular}{cccccccc}
\toprule
\bf Size & \bf Hidden Size & \bf GLU Size & \bf \#Heads & \bf \#Layers & \bf Batch Size & \bf \# Tokens & \bf Seq Length \\
\midrule
700M & 1536 & 4096 & 24 & 24 & 1M & 100B & 2048 \\
1.3B & 2048 & 5460 & 32 & 24 & 1M & 100B  & 2048  \\
3B & 3200 & 8640 & 32 & 26 & 1M & 100B  & 2048  \\
7B & 4096 & 11008 & 32 & 32 & 1M & 100B  & 2048  \\
\bottomrule
\end{tabular}
\caption{Model configurations for both \our{}, BitNet b1.58 and \llama{}.}
\end{table*}

\begin{table*}[h]
\centering
\begin{tabular}{lcccccc}
\toprule
\bf Model & \bf Size & \bf Learning Rate & \bf Weight Decay & \bf Warm-up & \bf Adam $\beta$ \\
\midrule
\multirow{4}{*}{\our{}} & 700M & $1.5\times10^{-3} \rightarrow 1\times10^{-3}$ & $0.1 \rightarrow 0$ & 375 & (0.9, 0.95) \\
& 1.3B & $1.2\times10^{-3} \rightarrow 8\times10^{-4}$ & $0.1 \rightarrow 0$ & 375 & (0.9, 0.95)  \\
& 3B &  $1.2\times10^{-3} \rightarrow 6.4\times10^{-4}$ & $0.1 \rightarrow 0$ & 375 & (0.9, 0.95) \\
& 7B &  $1\times10^{-3} \rightarrow 6\times10^{-4}$ & $0.1 \rightarrow 0$ & 375 & (0.9, 0.95) \\
\midrule
\multirow{3}{*}{\llama{}} & 700M & $2.5\times10^{-4}$ & 0.1 & 375 & (0.9, 0.95) \\
 & 1.3B & $2.0\times10^{-4}$ & 0.1 & 375 & (0.9, 0.95) \\
 & 3B &  $2.0\times10^{-4}$ & 0.1 & 375 & (0.9, 0.95) \\
 & 7B &  $1.5\times10^{-4}$ & 0.1 & 375 & (0.9, 0.95) \\
\bottomrule
\end{tabular}
\caption{Hyper-parameters for both \our{} and \llama{} training.}
\end{table*}

\end{document}